\pdfoutput=1
\documentclass{article} 
\usepackage{nips12submit_e,times}
\usepackage{url,epsfig}
\usepackage{hyperref}
\usepackage{tabularx}
\usepackage{multirow}
\usepackage{amsmath} 

\title{On Training Bi-directional Neural Network Language Model with Noise Contrastive Estimation}

\author{
Tianxing He \\
Shanghai Jiao Tong University \\
\texttt{cloudygoose@sjtu.edu.cn} \\
\And
Yu Zhang\\
MIT\\
\texttt{yzhang87@csail.mit.edu} \\
\And
Jasha Droppo \\
Microsoft Research \\
\texttt{jdroppo@microsoft.com} \\
\And
Kai Yu \\
Shanghai Jiao Tong University \\
\texttt{kai.yu@sjtu.edu.cn} \\
}

\nipsfinalcopy 

\begin{document}

\renewcommand{\arraystretch}{1.3}

\maketitle

\begin{abstract}
We propose to train bi-directional neural network language model(NNLM) with noise contrastive estimation(NCE). Experiments are conducted on a rescore task on the PTB data set. It is shown that NCE-trained bi-directional NNLM outperformed the one trained by conventional maximum likelihood training. But still(regretfully), it did not out-perform the baseline uni-directional NNLM.
\end{abstract}

\section{Introduction}
\label{intro}
Recent years have witnessed exciting performance improvements in the field of language modeling, largely due to introduction of a series of neural network language models(NNLM). Although the conventional back-off n-gram language model has been widely used in the automatic speech recognition (ASR) or machine translation(MT) community for its simplicity and effectiveness, it has long suffered from the \textit{curse-of-dimensionality} problem caused by huge number of possible word combinations in real-world text. Various smoothing techniques\cite{chen96} are proposed to address this issue but the improvements have been limited. Recently, neural network based language models have attracted great interest due to its effective encoding of word context history \cite{Bengio03aneural,Schwenk07,hnnlm05,improvenn10}.In neural network based language models, the word context is projected into a continuous space and the projection, represented by the transformation matrices in the neural network, are learned during training. The projected continuous word vectors are also referred to as {\em word embeddings}. With the continuous context representation, feed-forward neural network language models (FNNLM)\cite{Bengio03aneural,Schwenk07,hnnlm05,improvenn10,MnihHinton2007},  have achieved both better perplexity(PPL) and better word error rate (WER) when embedded into a real-world system.

Despite the benefits of effective context representation brought by word embeddings, FNNLM is still a short-span language model and not capable of utilizing long-term (e.g. context that is 5 or 6 words away) word history for the target word prediction. To address this issue, recurrent neural network language model (RNNLM), which introduces a recurrent connection in the hidden layer, is proposed to preserve long-term context. It has achieved significant performance gain on  perplexity and word error rate (WER) performance on various data sets \cite{rnnlm10,nnlmcompare13,lstmlm12,rnncache14,rnnrescore14,gpu14}, out-performing traditional back-off n-gram models and FNNLMs. 

However, RNN training generally suffered from the ``vanishing gradient'' problem\cite{Hochreiter01gradientflow}:the gradient flow will decay sharply through a non-linear operation. The LSTM\cite{lstm97} structure alleviates this problem by introducing a ``memory cell'' structure which allows the gradient to travel without being squashed by a non-linear operation. Also, it has a set of gates which enable the model to decide whether to memorize, forget, or output information. By introducing the LSTM strucutre into RNNLM\cite{lstmlm12}, LSTMLM is able to remember longer context information and gains more performance gain. It has also been shown that the dropout \cite{dropoutlstm14} can be used to regularize the LSTMLM. Inspired by its success, several variants of LSTM have been proposed, recently the gated recurrent unit(GRU)\cite{GRUvsLSTM14} is gaining increasing popularity becuase it has matching performance with LSTM but has simpler structure. More recently, \cite{memory15} has proposed to introduce the concept of memory into NNLM. By fetching memories from previous time, the model is able to ``explicitly'' utilizing long-term dependency without recurrence structure. 

While these research efforts have been focusing on better utilization of history information, it would be desirable if the model can utilize context information from both sides. In literature, very few attempts have been made to train a proper bi-directional neural network language model, even though bi-drectional NN has already been  successfully applied to other fields\cite{icmlctc14}. This is because the bi-directional model won't be by itself normalized because of the generative nature of language model, which makes the conventional maximum likelihood training framework improper for its training.

In this work, attempts have been made to train a bi-directional neural network language model with noise contrastive estimation, an alternative to maximum likelihood training which does not have the constrain that the model to be trained is inherently normalized. The rest of the paper is organized as follows: in section \ref{moti}, the motivation of this work is discussed, in section \ref{sec:form} the formulation of the model are elaborated in detail, implementation is covered in section \ref{sec:imple}, finally experiment results are shown in section \ref{sec:exp} and related works are discussed in section \ref{sec:relate}.

\section{Motivation}
\label{moti}
Statistical language models assign a probability $P(\mathcal{W})$ to a given sentence $\mathcal{W}=<w_1,w_2,...,w_n>$, which can be decomposed into a product of word-level probabilities using the rule of conditional probability:
\begin{equation}
P(\mathcal{W})=\Pi_{i} P(w_{i}|w_{1..i-1}) 
\label{eq:wcondi}
\end{equation}

Language models by this formulation predict the probability distribution of the next word given its former words(history). Since the prediction only depends on history information, in this work, this kind of model is denoted as uni-directional language model. All types of language model mentioned in section \ref{intro} fall into this category, but note that shot-span models like N-gram makes the "Markov Chain" assumption $P(w_{i}|w_{1..i-1}) \approx P(w_{i}|w_{i - N..i-1})$ to alleviate the data-sparsity problem.

For uni-directional language models, as long as each word-level probability is properly normalized, normalization is also guaranteed on sentence level:
\begin{equation}
\sum _ \mathcal{W} P_{LM}(\mathcal{W}) = 1
\label{eq:wnorm}
\end{equation}
This is the key reason why the "maximum likelihood estimation" training framework, which requires the model to be inherently probabilistic, has been successfully applied to the parameter estimation(training) of uni-directional language models. And recent years of research effort in the field of neural network language model has been focused on getting a better representation of history context using sophisticated recurrent neural network structures like LSTM\cite{lstmlm12}.

Unfortunately, while recently bi-directional neural network like BI-RNN or BI-LSTM has been successfully applied to many tasks, it is not trivial to apply this powerful model to language modeling, the main challenge is that the bi-directional information will break the sentence-level normalization\footnote{If some bi-directional model like $P(w_i|w_{1..i-1,i+1..N})$ is used as the word-level LM, equation \ref{eq:wcondi}, and hence equation \ref{eq:wnorm}
 won't hold any more.}, making the model no longer valid for the MLE training framework(please refer to section \ref{sec:form} for more details
).
 
In this work, noise contrastive estimation(NCE)\cite{ncejmlr12} is used to train a bi-directional neural network based LM, one big advantage of NCE over MLE is that it doesn't require the model to be self-normalized. This enables the utilization of bi-directional information for word-level scoring. Formulations of this work will be elaborated in the next section.

\section{Formulation}
\label{sec:form}
\subsection{Model Formulation}
In this work, $P(\mathcal{W})$ is the product of word-level scores(similar to uni-directional LM) and a learned normalization scalar $c$, required by the NCE framework to ensure normalization:
\begin{equation}
\begin{split}
f'(\mathcal{W})=\Pi_{i} f_i(\mathcal{W}) \\
P^{NCE}(\mathcal{W})=f'(\mathcal{W})exp(c)
\end{split}
\end{equation}
where $f_i$ the scoring given by a bi-directional neural network on each word index. And the "NCE" superscript for $P^{NCE}(\mathcal{W})$ is for indicating the normalization is induced by NCE training.

\begin{figure}
\centering
\includegraphics[bb = 80 490 530 750, clip=true, width=10cm]{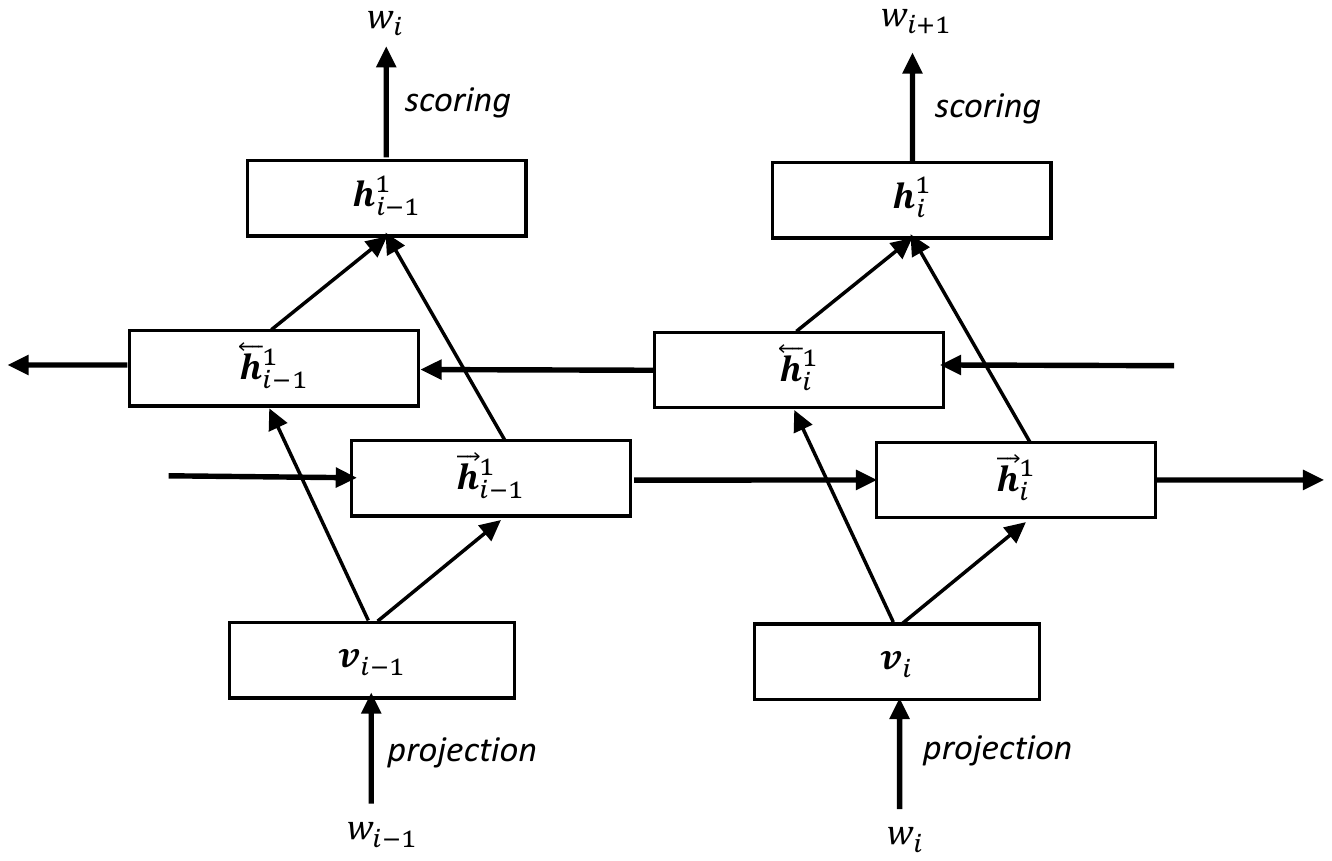}
\vspace{-0.1in}
\caption{\label{fig:birnn}Illustration of the network structure}
\vspace{-0.15in}
\end{figure}

In this work, the same bi-directional neural network structure that has been used in \cite{icmlctc14,bilm15} is applied, and is shown in figure \ref{fig:birnn} and formulated below(we are aware that other variants of BI-RNN exist\cite{birnng15}, but they are not fundamentally different with regard to this work):
\begin{equation}
\begin{split}
\textbf{v}_i&=\textbf{W}_{xh}\textbf{x}_i \\
\overrightarrow{\textbf{h}}^1_{i}&=g(\overrightarrow{\textbf{h}}^1_{i-1},\textbf{v}_i) \\
\overleftarrow{\textbf{h}}^1_{i}&=g(\overleftarrow{\textbf{h}}^1_{i+1},\textbf{v}_i) \\
\textbf{h}^1_i&= tanh(\textbf{W}^1_{hf}\overrightarrow{\textbf{h}}^1_{i} + \textbf{W}^1_{hr}\overleftarrow{\textbf{h}}^1_{i} + \textbf{b}^1) \\
\textbf{u}_i &= exp(\textbf{W}_{ho} \textbf{h}^1_i + \textbf{b}_o)
\end{split}
\end{equation}
where $\textbf{W}_{**}$ and $\textbf{b}_{*}$ are the transformation matrices and bias vector parameters in the neural network, and $\textbf{x}_i$ is the one-hot representation of $w_i$. Finally, $f_i(\mathcal{W})$ is obtained after a normalizing operation over the vocabulary(denoted as $\mathcal{V}$) on $\textbf{u}_i$:
\begin{equation}
f_i(\mathcal{W}) = \frac{\textbf{u}_i(w_i)}{\sum_{w_j \in V}\textbf{u}_i(w_j)}
\end{equation}
Note that the word-level normalization is not needed in this work($\textbf{u}_i(w_i)$ can be used directly as $f_i(\mathcal{W})$), but experiments show that reserving the word-level normalization will give better results.

In this work, gated recurrent unit is used as the recurrent structure $\textbf{h}_{t}=g(\textbf{h}_{t-1},\textbf{v}_t)$ because it is faster, causes less memory and has matching performance with the LSTM structure\cite{GRUvsLSTM14}. So our NN model is denoted \textbf{BI-GRULM} and the formulation is put below:
\begin{equation}
\begin{split}
\textbf{z}_t&=\sigma(\textbf{W}_{hz}\textbf{h}_{t-1}+\textbf{W}_{xz}\textbf{v}_{t}+\textbf{b}_z) \\
\textbf{r}_t&=\sigma(\textbf{W}_{hr}\textbf{h}_{t-1}+\textbf{W}_{xr}\textbf{v}_{t}+\textbf{b}_r) \\
\tilde{\textbf{h}}_t&=tanh(\textbf{W}_{h1}(\textbf{r}_t \ast \textbf{h}_{t-1})+\textbf{W}_{h2}\textbf{v}_{t} + \textbf{b}_h)\\
\textbf{h}_t&= (1-\textbf{z}_t) \ast \textbf{h}_{t-1}  + \textbf{z}_t \ast \tilde{\textbf{h}}_t
\end{split}
\end{equation}
where $\sigma$ is the sigmoid function $\sigma(x)=\frac{1}{1+e^{-x}}$ and $\ast$ is element-wise multiplication, and note that a different set of parameter is used for forward and backward connections in the bi-directional neural network.

Finally, we also write down the formulations of a one-layer uni-directional GRULM(UNI-GRULM) here since it will be used as baseline model:
\begin{equation}
\begin{split}
\textbf{h}^1_i=&g(\textbf{h}^1_{i-1},\textbf{W}_{xh}\textbf{x}_i) \\
\textbf{u}_i=&exp(\textbf{W}_{ho}\textbf{h}^1_i+\textbf{b}_o)
\end{split}
\end{equation}
Note that other than being uni-directional, the only other difference between these two models is the normalization scalar $c$. And in this work, the dropout operation is applied on $\textbf{h}^1_i$ for both models.

\subsection{Training of bi-directional NNLM}
As stressed in section \ref{moti}, the MLE framework is not suitable for training bi-directional NNLM, still, in this work MLE training is tried as a baseline experiment. Denoting the the data distribution as $P_{data}(\mathcal{W})$, the MLE objective function is formulated as below:
\begin{equation}
J_{MLE}(\theta)=E_{P_{data}(\mathcal{W})}[logf_{\theta}'(\mathcal{W})]
\end{equation}
Note that here the normalization scalar $c$ does not exist in the model.

In this work, noise contrastive estimation\cite{ncejmlr12} is applied to train the bi-directional NNLM. NCE introduces a noise distribution $P_{noise}(\mathcal{W})$ into training and a "to-be-learned" normalization scalar $c$ into the model, and its basic idea is that instead of maximizing the likelihood of the data samples, the model is asked to discriminative samples from the data distribution against samples from the noise distribution:
\begin{equation}
\begin{split}
J_{NCE}(\theta)=E_{P_{data}(\mathcal{W})}[logP(D=&1|\mathcal{W};\theta)] + kE_{P_{noise}(\mathcal{W})}[logP(D=0|\mathcal{W};\theta)] \\
P(D=1|\mathcal{W};\theta)&=\frac{P^{NCE}_{\theta}(\mathcal{W})}{P^{NCE}_{\theta}(\mathcal{W}) + kP_{noise}(\mathcal{W})} \\
P(D=0|\mathcal{W};\theta)&=\frac{kP_{noise}(\mathcal{W})}{P^{NCE}_{\theta}(\mathcal{W}) + kP_{noise}(\mathcal{W})} \\
\end{split}
\end{equation}
assuming a noise ratio of $k$.

And the gradients are:
\begin{equation}
\begin{split}
\frac{\partial logP(D=1|\mathcal{W};\theta)}{\partial \theta} = \frac{kP_{noise}(\mathcal{W})}{P^{NCE}_{\theta}(\mathcal{W}) + kP_{noise}(\mathcal{W})} \frac{\partial log P^{NCE}_{\theta} (\mathcal{W})}{\partial \theta} \\
\frac{\partial logP(D=0|\mathcal{W};\theta)}{\partial \theta} = \frac{-P^{NCE}_{\theta}(\mathcal{W})}{P^{NCE}_{\theta}(\mathcal{W}) + kP_{noise}(\mathcal{W})} \frac{\partial log P^{NCE}_{\theta} (\mathcal{W})}{\partial \theta}
\end{split}
\end{equation}
 
For NCE to get good performance, a noise distribution that is close to the real data distribution is preferred, so in our case it is natural to use a good uni-directional LM as the noise distribution. In this work, N-gram LM is used as the noise distribution since it is efficient to sample from.  Details about implementation and training process will be covered in section \ref{sec:imple}.

\section{Training and Implementation details}
\label{sec:imple}
\begin{figure}
\centering
\includegraphics[bb = 80 625 530 770, clip=true, width=10cm]{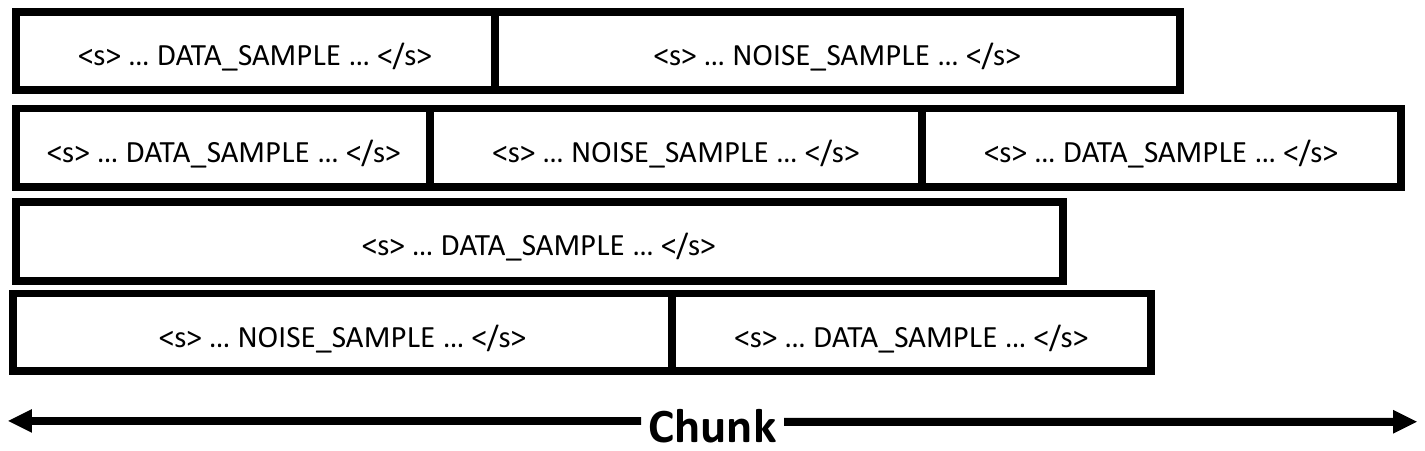}
\caption{\label{fig:chunk_imple}Illustration of the parallel training implementation}
\end{figure}

Mini-batch based stochastic gradient descent(SGD) is used to train bi-directional NNLM in this work. The training process is very similar to \cite{gpu14}, but several changes need to be made for the sentence-level bi-directional NNLM training. Since NN training in this work is sentence-level, data(consisted of real data samples and noise model samples) are processed in chunks, illustrated in figure \ref{fig:chunk_imple}. Moreover, a batch of data streams is processed together to utilize the computing power of GPU. It is relatively easy to realize this training process of BI-RNN with the help of neural network training tool-kits like CNTK\cite{cntk14}. In this work, the chunk size is set to 90(which is larger than the longest sentence in the ptb data-set) and the batch size is set to 64.

An validation-based learning strategy is used, the learning rate is fixed to a large value at first, and start halving at a rate of 0.6 when no significant improvement on the validation data is observed. And the training is stopped when that happens again. Further, a L2 regularization with coefficient 1e-5 is used.

Finally, the SRILM\cite{srilm02} Toolkit is used for N-gram LM training in this work. In our training the N-GRAM noise is generated on-the-fly so noise samples won't be the same between iterations.

\section{Experiments}
\label{sec:exp}
\subsection{Datasets}
\label{sec:data}
In this section, results of experiments designed to test the performance of the proposed bi-directional NNLM trained by NCE. Since the training process is very time-costly when the noise ratio $k$ is large(in our training framework, it will cost at least $k$ times the time for training the baseline UNI-GRULM model), we confined our experiments to the Penn Treebank portion(PTB) of the WSJ corpus, which is publicly available and has been used extensively in LM community. There are 930k tokens, 74k tokens, 82k tokens for training, validation and testing, respectively, and the vocabulary size is 10k. 

Further, since there is no guarantee that the trained model will be properly normalized, the evaluation of perplexity(PPL), which is the most conventional evaluation for LM, can no longer be applied. Instead, we need to resort to some discriminative task in which the LM is asked to tell "good" sentence from "bad" sentences, like its application in decoding or rescoring in systems like speech recognition or machine translation. But still, we want the training corpus and vocabulary size to be small enough, which will enable us to try a large noise ratio $k$, since sentence-level sampling is considered in this work, it is expected that $k$ needs to be large enough for the training to work.

In light of the above concerns, a rescoring task is created directly on the PTB dataset, denoted as \textbf{ptb-rescore}\footnote{This test set and the scripts for reproducing the N-gram baseline are available at \url{https://bitbucket.org/cloudygoose/ptb_rescore}}. In this test, random small errors are introduced to each sentence of the original test corpus of the PTB dataset, and the LM is then asked to recognize the original sentence from the tampered ones by assigning it the highest score. In this work, three types of error, namely \textbf{substitution}, \textbf{deletion} and \textbf{insertion}, are generated. For each error type, 9 decoys(one decoy only has one error) are generated for each test sentence, constituting three test sets. So a uniform guess will have an accuracy of $10\%$. Further, a mixed set where each decoy can be of any of the three types of error is also added, denoted as test \textbf{sdi}. Some examples are shown in table \ref{tab:exptb}. Note that in this test set all random number(for the position or new word index) are drawn from a uniform distribution, and the \textbf{s-test} set is similar to the MSR sentence completion task \cite{msrsc11}.

\begin{table}
\centering
\begin{tabular}{ c c } 
 \hline
 \textit{original} & no it was n't black monday \\
 \hline
 \textit{s-error} & no it was n't black \textbf{revoke} \\
 \textit{d-error} & no it was n't monday \\
 \textit{i-error} & no it \textbf{cracks} was n't black monday \\
\hline
\end{tabular}
\caption{\label{tab:exptb}Examples of decoys in the \textbf{ptb-resocre} test set}
\end{table}

\subsection{Pseudo-PPL Test}
Although perplexity can not be used to evaluate bi-directional NNLM, it is still interesting what PPL the trained model will assign to the test sentences. Besides the original test set for the \textbf{PTB} data, two additional text are generated, one is sentences sampled from the 4-GRAM baseline model(denoted as \textbf{4gram-text}), the other one is sentences sampled from a completely uniform distribution(denoted as \textbf{uniform-text}). All three sets have around 4,000 sentences. A well-behaved LM is expected to assign lowest PPL to the first set, relatively low PPL to the second, and very high(bad) PPL to the last one. The results are shown in table \ref{tab:ptbppl}.

\begin{table}[htbp!]
\centering
\begin{tabular}{ c|c c c } 
 \hline
  \multirow{2}{*}{Model} & \multicolumn{3}{c}{Pseudo-PPL} \\ 
  & \textbf{test-ptb} & \textbf{4gram-text} & \textbf{uniform-text} \\
\hline
UNI-GRULM & 103.7 & 431.0 & 91935.7 \\
BI-GRULM(MLE) & 1.12 & 1.16 & 3.358 \\
BI-GRULM(NCE with noise ratio $10$) & 15.5 & 3846.4 & 99565.4 \\
\hline
\end{tabular}
\caption{\label{tab:ptbppl}Pseudo-PPL result of different trained LMs on three test sets.}
\end{table}

It is shown that the BI-GRULM(detailed configuration will be discussed in section \ref{sec:ptbrescore}) trained with NCE has similiar behavior to the baseline uni-directional model, meaning that NCE is helping the model with sentence-level normalization. On the contrary, BI-GRULM trained with MLE is assigning extremely low PPL to every test set, indicating that the model is not properly normalized. But surprisingly, the relative order of PPL from MLE-trained BI-GRULM is correct.

\subsection{Evaluation on the ptb-rescore task}
\label{sec:ptbrescore}
In this section accuracy results on the \textbf{ptb-rescore} task is presented. Three models are trained to be baseline models: 4-GRAM, UNI-GRULM, and BI-GRULM trained by MLE. Note that unless otherwise mentioned, all GRULMs trianed in the work has 300 neurons on hidden layer and only one layer(in the BI-GRULM case, one layer means one forward layer and one backward layer) is used. This setting is chosen for the reason that adding more neurons or more layers give no significant on the test PPL for the baseline UNI-GRULM model. Through training, a dropout rate of $50\%$ is applied for the UNI-GRULM, but no dropout is applied for the reported experiments for the BI-GRULM because it is found that dropout won't give performance gain in that case.

The baseline results are shown in the upper part of table \ref{tab:ptbresexp}. Overally, the UNI-GRULM model gives the best performance, as expected. An interesting observation is that all model have extremely poor performance on the \textbf{test-d} set. This behavior, however, is not so surprising since the LM score of a sentence is afterall a product of word-level probabilities, so decoys with one less word will have big advantage. It is found that this problem can be alleviated by a \textbf{length-norm} trick:
\begin{equation}
score_{length-norm}(\mathcal{W})=\frac{score(\mathcal{W})}{l}=\frac{logf(\mathcal{W})}{l}=\frac{\sum^l_ilogf_i(\mathcal{W})}{l}
\end{equation}
assuming sentence $\mathcal{W}$ is of length $l$(including the sentence-end token). Note that this trick is equivalent to ranking the sentences using PPL instead of sentence-level log likelihood and it will do harm to the performance on the \textbf{test-i} set, although not large.

\begin{table}[htbp!]
\centering
\begin{tabular}{ c | c|c c c c} 
 \hline
  \multirow{2}{*}{Model} & noise & \multicolumn{4}{c}{Accuracy(\%)/Accuracy after \textbf{length-norm}(\%)} \\ 
 & ratio & \textbf{test-s} & \textbf{test-d} & \textbf{test-i} & \textbf{test-sdi} \\
\hline
4-GRAM & - & 75.4/n75.4 & 3.2/n12.7 & \textbf{100}/n\textbf{98.2} & 13.4/n40.8 \\
UNI-GRULM & - & \textbf{80.6}/n\textbf{80.6} & \textbf{3.9}/n\textbf{21.8} & 99.9/n96.9 & \textbf{20.2}/n\textbf{60.9} \\
\hline
BI-GRULM(MLE) & - & 50.0/n50.0 & 0.31/n21.9 & 95.3/n31.5 & 6.8/n27.1 \\
\hline
 & 1 & 31.9/n31.9 & 3.9/n12.8 & 67.4/n53.0 & 10.9/n17.8 \\
BI-GRULM & 10 & 39.9/n39.9 & 8.8/n19.4 & 61.8/n48.8 & 20.5/n26.2 \\
(NCE) & 20 & 39.2/n39.2 & \textbf{11.0}/n\textbf{21.6} & 59.1/n45.3 & \textbf{21.0}/n26.3 \\
 & 50 & 48.4/n48.4 & 6.8/n19.8 & 74.2/n54.9 & 18.1/n29.0 \\
 & 100 & \textbf{55.7}/n\textbf{55.7} & 0.5/n13.4 & \textbf{98.6}/n\textbf{80.4} & 10.3/n\textbf{34.5} \\
\hline
\end{tabular}
\caption{\label{tab:ptbresexp}Accuracy result of BI-GRULM models trained by NCE.}
\end{table}

Results of BI-GRULM trained by NCE are shown in in the lower part of table \ref{tab:ptbresexp}, it is observed that the \textbf{length-norm} trick can also help in this case, and the overall performance is improving with larger and larger noise ratio, however, it became unaffordable for us to run experiments with ratio larger than 100. One strange observation is that performance on the \textbf{test-d} set degrades with larger noise ratio, and this causes performance on the \textbf{test-sdi} to become worse. Also, comparing with the BI-GRULM(MLE) result, BI-GRULMs trained by NCE with a large noise ratio have overally better performance, indicating that NCE has the potential to utilize to power of BI-GRULM structure more properly. 

Unfortunately, the proposed model failed to out-perform the best UNI-GRULM baseline model on every test set. Results on the \textbf{test-s} set show that improvement can only be obtained by growing the noise ratio exponentially, this matches our concern in section \ref{sec:data}, the sentence-level sampling space may be too sparse for our sampling to properly cover.

\section{Related work}
\label{sec:relate}
In \cite{bilm15}, bi-directional LSTMLM is trained with MLE and tested by LM rescoring in an ASR task. However, no improvement is observed over the uni-directional baseline model. On the other hand, NCE has been used in uni-directional LM training both for FNNLM\cite{ncelm12} and RNNLM\cite{ncernnlm15}, the main goal was to speed-up the training and evaluation of these two models because under NCE training the final softmax operation on the output layer is no longer necessary. Note that different from these two work, NCE is applied on the sentence level in this work.

\section{Conclusion}
In this work noise contrastive estimation is used to train a bi-directional neural network language model. Experiments are conducted on a rescore task on the PTB data set. It is shown that NCE-trained bi-directional NNLM outperformed the one trained by conventional maximum likelihood training. But still, it did not out-perform the baseline uni-directional NNLM. The key reason maybe that the sentence-level sampling space is too sparse for our sampling to cover.

\section{Acknowledgements}
The authors want to thank Abdelrahman Mohamed, Kaisheng Yao, Geoffrey Zewig, Dong Yu, Mike Seltzer, and Da Zheng for valuable discussions.

\bibliography{icml16_htx}
\bibliographystyle{IEEEbib}


\end{document}